\documentclass{article}

\usepackage{microtype}
\usepackage{graphicx}
\usepackage{multicol}
\usepackage{amsfonts}
\usepackage{float} 
\usepackage{subfigure}
\usepackage{enumitem}%
\usepackage{amssymb}
\usepackage{times, graphicx, enumitem}
\usepackage[ruled,algo2e]{algorithm2e}
\usepackage{scrextend}
\usepackage{booktabs}
\usepackage{hyperref}

\newcommand{\Cc}{\mathcal{C}}
\newcommand{\Ii}{\mathcal{I}}
\newcommand{\Rr}{\mathcal{R}}

\newcommand{\Aa}{\mathcal{A}}

\usepackage{xcolor}
\definecolor{Blue}{rgb}{0.0, 0.48, 0.65}
\definecolor{Green}{rgb}{0.0, 0.56, 0.0}
\definecolor{Pink}{rgb}{0.93, 0.51, 0.93}

\usepackage[accepted]{style}

\begin{document}

\twocolumn[
\XXXtitle{Archtree: on-the-fly tree-structured exploration for latency-aware pruning of deep neural networks}

\XXXsetsymbol{equal}{*}

\begin{XXXauthorlist} 
\XXXauthor{Rémi Ouazan Reboul}{datakalab}
\XXXauthor{Edouard Yvinec}{datakalab,isirupmc}
\XXXauthor{Arnaud Dapogny}{datakalab}
\XXXauthor{Kevin Bailly}{datakalab,isirupmc}% equal to goo ed
\end{XXXauthorlist}

\XXXaffiliation{datakalab}{Datakalab, 114 boulevard Malesherbes, 75017 Paris, France}
\XXXaffiliation{isirupmc}{Sorbonne Université, CNRS, ISIR, f-75005, 4 Place Jussieu 75005 Paris, France}

\XXXcorrespondingauthor{Rémi Ouazan Reboul}{remi.pierre\_o@orange.fr}

% You may provide any keywords that you
% find helpful for describing your paper; these are used to populate
% the "keywords" metadata in the PDF but will not be shown in the document
\XXXkeywords{pruning, latency, deep neural networks, hardware}

\vskip 0.3in

\begin{abstract}

Deep neural networks (DNNs) have become ubiquitous in addressing a number of problems, particularly in computer vision. However, DNN inference is computationally intensive, which can be prohibitive e.g. when considering edge devices. To solve this problem, a popular solution is DNN pruning, and more so structured pruning, where coherent computational blocks (e.g. channels for convolutional networks) are removed: as an exhaustive search of the space of pruned sub-models is intractable in practice, channels are typically removed iteratively based on an importance estimation heuristic. Recently, promising latency-aware pruning methods were proposed, where channels are removed until the network reaches a target budget of wall-clock latency pre-emptively estimated on specific hardware. In this paper, we present Archtree, a novel method for latency-driven structured pruning of DNNs. Archtree explores multiple candidate pruned sub-models in parallel in a tree-like fashion, allowing for a better exploration of the search space. Furthermore, it involves on-the-fly latency estimation on the target hardware, accounting for closer latencies as compared to the specified budget. Empirical results on several DNN architectures and target hardware show that Archtree better preserves the original model accuracy while better fitting the latency budget as compared to existing state-of-the-art methods.

\end{abstract}
]

\printAffiliationsAndNotice{}  

\section{Introduction}\label{sec:intro}
Deep Neural networks (DNNs) are the cornerstone of many recent advances, particularly in computer vision. However, their adaptability and performance come at the cost of designing ever larger models with many parameters, hence making inference a slow process. This is especially true when it comes to running models on edge devices or consumer GPUs. There exists numerous techniques to speed up inference, among which is DNN pruning. Such methods generally aim at compressing the model, whether it be in terms of memory footprint, direct latency, or throughput gains. The field of DNN pruning can be divided into several subdomains, depending on the granularity of the pruning. As such, unstructured pruning \cite{lin2020dynamic,park2020lookahead,lee2019signal} consists in sparsifying the weight tensors without seeking to enforce a particular pattern, whereas semi-structured pruning \cite{holmes2021nxmtransformer,yvinec2022red++} enforces more or less constrained patterns in the sparsity, making them easier to leverage. Last but not least, structured pruning \cite{liebenwein2019provable,li2016pruning,he2018soft,wang2021convolutional,redwaraids} consists in removing coherent computational blocks (e.g. whole neurons or channels, when applied to convolutional networks) from the DNN.

\begin{figure*}[!t]
    \centering
    \includegraphics[width = 0.95\linewidth]{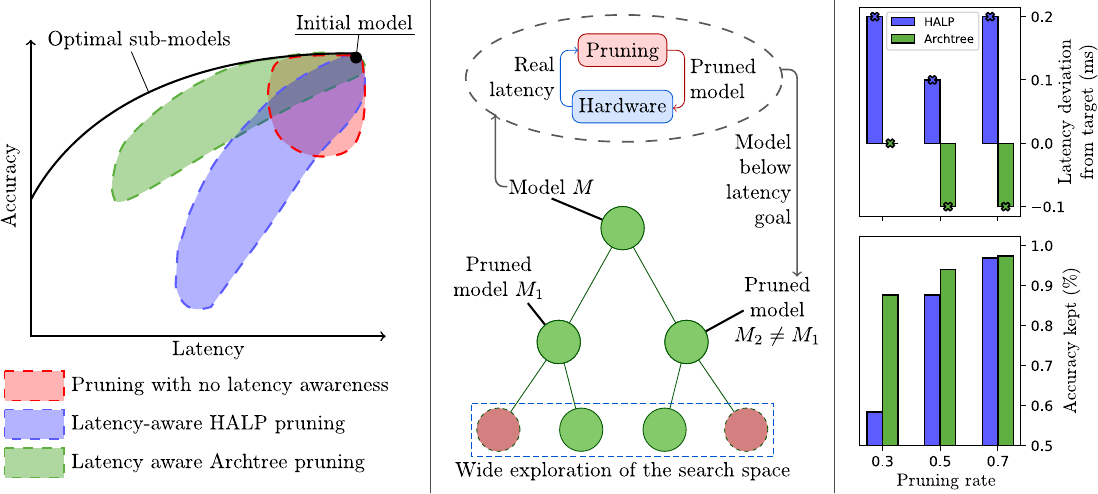}
    \caption{Towards bridging the gap with the intractable exhaustive search for the best pruned sub-model, latency-aware methods such as HALP \cite{halp} and Archtree use latency estimation to find better trade-offs (left). Archtree improves over HALP both in terms of accuracy of the pruned models (right-lower for a ResNet-8 on STM32 hardware) and how close it fits the latency goal (right-upper). This is due to a more efficient tree-structured search of the search space as well as on-the-fly \textit{in situ} latency estimation of the sub-models (middle). Numbers for the rightmost figure are from Table \ref{tab:comp8}.}
    \label{fig:main_fig}
\end{figure*}

If the goal is to specifically reduce the latency of the model, the latter category of DNN pruning methods is particularly interesting, since removing whole channels translates rather straightforwardly \cite{liu2021latency} into latency gains at inference time. However, modern DNNs usually have a very large number of parameters (from $100k$ to several billions), making an exhaustive search for the best sub-models (where channels of several layers have been pruned, see black line on Figure \ref{fig:main_fig}-left plot) intractable in practice. To address this problem, most structured pruning methods rely on heuristics to estimate the impact of removing neurons on the final accuracy. For instance, \citet{peng2019collaborative} modelize the inter-channel dependencies as well as the joint impact of pruned and kept channels on the final loss. \citet{guo2021gdp} add learnable gate parameters to zero-out particular channels after training is done. In the work of \citet{yu2022hessian}, the authors propose to use the hessian of the loss function w.r.t. the network's weights to estimate the less sensitive channels. Lastly, \citet{yvinec2022singe} argue that all the gradient-based channel importance measurements are intrinsically local, and borrow techniques from the field of visual explanation in DNNs to derive a novel, less local, integrated gradient importance criterion to remove the least important channels, outperforming the current state-of-the-art structured pruning methods.

A common drawback of the aforementioned methods is that while removing channels has a direct impact on the latency of the pruned network on most pieces of hardware, there is no guarantee that these methods find good accuracy \textit{v.s.} latency trade-offs in practice (red blob in Figure \ref{fig:main_fig}-left plot). For instance, removing channels from e.g. the first layers in the network (where feature maps are usually bigger) potentially bears more impact on the latency of the model than those towards the end layers. Therefore, we argue that, to find better such trade-offs, one shall use the sub-model latency-possibly estimated directly on a target hardware and inference engine (as relative latency among DNN's layers may vary depending on both \cite{nnmeter}) to drive the pruning process. As such, recently, \citep{halp} introduced HALP, a latency-oriented structured pruning algorithm. Similarly to other structured pruning methods \cite{yu2022hessian, yvinec2022singe}, HALP identifies expendable neurons by using an importance criterion. It then predicts the latency of possible pruned sub-models using a lookup table and formulates structured pruning as a knapsack problem, where each neuron is assigned a value (its importance) and a weight (its latency). The problem then becomes to maximize the value of a knapsack with a maximal weight capacity, \textit{i.e.} a maximal latency.

By doing so, HALP \cite{halp} allows finding better accuracy \textit{i.e.} latency trade-offs (as illustrated by the blue blob in Figure \ref{fig:main_fig}-left plot). Nonetheless, this approach bears some limitations: first, it relies on building a lookup table to estimate the latency. This lookup table, however, is constructed on single layers extracted in a vacuum rather than the whole network, thus overlooking the effects of serialization, parallelization, or hardware-rooted side effects such as memory transfers or caching. Hence, in practice, models pruned with HALP have the tendency to not closely match the latency goal. Second, akin to greedy search, HALP only maintains a single best candidate sub-model during the whole pruning process, thus making it less likely to find one that retains high accuracy with low latency. 

%In this paper, our scope is to structurally prune the model in order to reduce its latency, as it is the best-suited for real-time inference on edge devices. Reducing latency increases the frames per second of the model, and structured pruning has the added benefit of reducing the model's memory footprint. Plus, structured pruning is the only kind of pruning that is leverageable on any hardware.

%The issue with latency-oriented structured pruning is twofold. First, structured pruning is more restrictive than non-structured pruning. "Useful" and "useless" neurons may be located in the same structure, meaning both have to be pruned at the same time, which is not an issue in non-structured pruning.

%Then, the way it groups neurons and their importance unnecessarily hinders pruning by making some pruned sub-models unattainable. 

To overcome these issues, in this paper, we propose Archtree, a novel latency-aware structured pruning method. Archtree involves tree-structured exploration of the search space of the pruned sub-models, as well as on-the-fly \textit{in situ} latency estimation on the target hardware, as illustrated in Figure \ref{fig:main_fig}-middle plot. Compared with HALP, Archtree allows to more closely fit the latency goal (Figure \ref{fig:main_fig}-upper right bar plot) and better-preserved accuracies at every pruning goal (Figure \ref{fig:main_fig}-lower right bar plot), thus overall better trade-offs (green blob on Figure \ref{fig:main_fig}-left plot). In summary, the contributions of this paper are:
\begin{itemize}
    \item A tree-structured exploration by maintaining several candidates pruned sub-models in parallel which, akin to beam search, allows better exploration of the search space and results in higher accuracies.
    \item An on-the-fly \textit{in situ} latency estimation of the whole sub-models on target hardware that allows to more closely fit the target latency budget.
\end{itemize}

Furthermore, we experimentally show that the proposed Archtree significantly outperforms existing baselines on several benchmarks including different DNN architectures, target pieces of hardware, and latency goals.

\section{Motivation}

\begin{figure}[!t]
    \centering
    \includegraphics[width=0.9\linewidth]{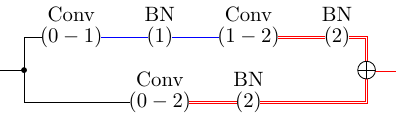}
    \caption{A ResNet basic block with its channel groups. We consider the input tensor to be part of channel group 0. Group 0 is defined because of shared inputs, group 1 is due to sequentiality and group 2 is caused by an addition node.}
    \label{fig:basic}
\end{figure}

Consider a pre-trained network $F(x): x \rightarrow f_L \circ f_{L-1} \circ \dots \circ f_1 (x)$ written as a composition of $L$ layers. If we prune a layer $l$ in a structured fashion, by removing one of its input or output channel, we may also need to prune other layers inside the network due to the sequentiality of the network (e.g. the presence of skip connections or add nodes). For this reason, we introduce the notion of \textit{channel group}: layers that are part of the same channel group are pruned together, in order to keep a coherent number of channels across the channel group. As an example, in figure \ref{fig:basic}, we show channel groups in a ResNet basic block. The output of the main and skip connections are added, which implies that their channels should be pruned simultaneously.

Thus, the network $F$ can be broken down into $N$ channel groups. This representation is more convenient from the perspective of a structured pruning method. We write $\Cc_n$ the number of channels in a channel group $n$, and $\Cc = \left( \Cc_1, \dots, \Cc_N \right)$ the vector containing the number of channels in each channel group. Structured pruning is equivalent to first choosing a new vector $\Cc'$ component-wise lesser or equal to $\Cc$ and then choosing which channel remains in each channel group. Since the latency of a model only depends on $\Cc$, we know the latency of the pruned model even before we select which channels remain in it. Then, by selecting the right channels to keep, we can maximize the accuracy of the newly pruned model. 

Similarly to \citet{halp}, our goal is to get a pruned model with maximal accuracy while having a latency below a preset latency goal $\tau_s$. We propose to represent the search space of possible pruned models, defined by their latency $\Cc'$ and accuracy, as a tree of candidate architectures, dubbed Archtree. The construction process of Archtree comprises two steps, illustrated in Figure \ref{fig:archtree}. First, we generate a set of architectures that reach a target intermediate latency goal, measured on the target hardware. Second, we eliminate architectures based on their estimated importance loss (a proxy for accuracy) and fine-tune the most promising candidates similarly to \citep{renda2020comparing}. This process is repeated until convergence to the target latency goal $\tau_s$.

In the following section, we detail the proposed approach, which relies on two crucial components: the on-device measured latency, and the importance.

% Similarly to \edouard{citer}, we prune the model progressively, maintaining a minimal accuracy through the pruning process.  Moreover, after each pruning, we train the model on a few batches in order to avoid catastrophic drops in the accuracy, of which the model could not recover from without a complete re-training. The pruning process is guided by two variables: the latency of possible pruned sub-model and the importance of each channel. We will first see how to measure importance (\ref{ssec:importance}). Then we will build a tree of pruned architectures (Archtree) (\ref{ssec:archtree}). It will structure how we explore possible sub-models and bnchmark their latency. Finally, we will see how to reduce the cost of navigating the Archtree with minimal impact (\ref{ssec:efficient}). \remi{je garde la dernière phrase? }

\begin{algorithm*}[ht]
\vspace{2pt}
\KwIn{
    \begin{multicols}{2}
    \begin{itemize}[noitemsep,topsep=1pt]
        \item a model $M$ with a latency $\tau_0$
        \item a latency goal $\tau_s < \tau_0 $
        \item a number of pruning steps $s \geq 1$ 
        \item a number of active nodes $A \geq 1$
        \item a set $\mathcal{D}$ of fine-tuning examples
    \end{itemize}
    \end{multicols} \vspace{-8pt}
}\KwOut{$A$ models $M_1, \dots, M_A$ with latencies $\tau^*_1, \dots, \tau^*_A$ below $\tau_s$.}
\vspace{5pt}

\Begin{
\DontPrintSemicolon $~~$\newline 
    Create root node $a_{root}$ associated with $M$ \newline 
    Measure its latency $\tau_0 = $ latency$(a_{root})$ \newline 
    Create the set of alive nodes $\Aa \leftarrow \lbrace a_{root} \rbrace$ \newline
    \rule{0.4\textwidth}{.4pt}\textbf{PRUNING PHASE}\rule{0.4\textwidth}{.4pt}\;
    
    \For{$i\leftarrow 1$ \KwTo $s$}{ $~$
        Set latency goal $\tau_i \leftarrow \frac{ \left( s - i \right) \tau_0 + i \tau_s }{s}$\newline 
        \For{$a \in \Aa$}{ $~$
            \textbf{Train} Finetune $a$'s associated model $M_a$ for a small number of iterations on $\mathcal{D}$ and gather importances \newline
            \textbf{Blossom} Prune each of $M_a$ channels group to create the children of $a$ with latency below $\tau_i$ \newline
            \textbf{Uniqueness} Delete children of $a$  that already exist (active or not) \newline
            \textbf{Loss} Calculate the importance loss $\Delta \mathcal{I}$ of each child w.r.t. the parent model $M_a$\newline
            \textbf{Death} Deactivate node $a$ (so $a$ is no longer in $\Aa$)
        } $~$
        \textbf{Importance filter} Shrink $\Aa$  by only keeping the $A$ most important nodes of $\Aa$ \newline
    }
    
    \rule{0.39\textwidth}{.4pt}\textbf{FINETUNE PHASE}\rule{0.39\textwidth}{.4pt} \newline
    Order $\left( M_a \right)_{a \in \Aa} $ so $M_1$ has the highest validation accuracy and $M_A$ the lowest\newline
    \For{$i\leftarrow 1$ \KwTo $A$}{ $~$
        \textbf{Finetune} Finetune the associated model $M_i$ on $\mathcal{D}$\newline
        \textbf{Benchmark} Measure the latency $\tau^*_i$ of the model 
    }
}
\caption{\bf Archtree} \label{ArchtreeAlgorithm}
\end{algorithm*}

\section{Methodology overview}\label{ssec:methodo}

\subsection{Computing importance}\label{ssec:importance}

At each pruning step, we need to compare the $N$ channel groups to remove channels from the least important ones. Below, we describe how we compute importance over \textit{weights} in $F$ and aggregate these measurements at the channel group level.

\paragraph{Weight importance criterion:} 

%Let's consider a channel group $C_k$ spanning across multiple layers $\{l_1, \dots, l_{L_k}\}$. Let $l$ denote one of these layers and $w_l \in \mathbb{R}^{o \times i \times k}$, with $o, i, k$ respectively indicating the output and input dimensions, as well as the kernel size ($k=1$ for dense layers, $k=k_k \times k_w$ for $k_k \times k_w$ convolutional kernels).

Let's consider a set of fine-tuning examples $\mathcal{D} = \{(x_j,y_j)\}_j$, and $l$ a prunable (\textit{i.e.} a dense or convolutional) layer with weights $W_l \in \mathbb{R}^{o \times i \times k}$, with $o, i, k$ respectively indicating the output and input dimensions, as well as the kernel size ($k=1$ for dense layers, $k=k_k \times k_w$ for $k_k \times k_w$ convolutional kernels). The importance of weight $W$ is defined by:
\begin{equation}\label{eq:wimp}
    I \triangleq \sum_j\left| W \cdot \frac{\partial \mathcal{L}(F(x_j),y_j)}{\partial W} \right|
\end{equation}
where $\mathcal{L}(F(x_j),y_j)$ denotes the final loss function (e.g. traditional softmax cross-entropy loss). This criterion outputs an importance tensor $I_l \in \mathbb{R}^{o \times i \times k}$. Intuitively, a weight is deemed important if small variations of this weight cause large perturbations in the loss function. This criterion offers the advantage of being easy to compute and relatively efficient, as pointed out in \cite{yvinec2022singe}.

\paragraph{Spatial reduction:} the spatial reduction step $\Rr_s$ aims at converting $I_l$ to an $o \times i$ layer importance matrix. If layer $l$ is dense, $I_l$ already has the desired format, hence $\Rr_s$ is simply an identity function. If layer $l$ is convolutional, $\Rr_s$ is either a sum, a mean or a max. These options will be discussed in Section \ref{exp:ablations}.

\paragraph{Neural reduction:} similarly, the layer-wise spatially reduced importance matrices $\Rr_s(I_l) \in \mathbb{R}^{o\times i}$ can be reduced row-wise (resp. column-wise) to get an $o$-length (resp. $i$-length) vector in order to compute the importance of the output (resp. input) channel group. This reduction, denoted $\Rr_n$, can consist of either sum, mean or max.

\paragraph{Channel group reduction:} lastly, since a channel group $n$ spans across multiple layers, we need to combine the importance vectors obtained from each layer. This reduction, denoted $\Rr_c$, summarizes all vectors $\Rr_n(\Rr_s(I_l))$ for all layers $l$ belonging to this channel group, into one $\Cc_n$-dimensional importance vector $\Ii_n$.

%This reduction takes an arbitrary number of $\Cc_k$-vectors and outputs one. We note $\Rr_c: \RR^{\Cc_k \times \cdots \times \Cc_k} \rightarrow \RR^{\Cc_k}$ this reduction, called channel group reduction. This means that the importance of channel group $k$ is defined by:

%where $(L)_k$ is the subset of layers where $k$ is an input or output channel group.

The importance criterion is the metric that determines which channels to remove \textit{inside} a channel group, with the convention that we always remove the least important channel first. Additionally, importance is used in the Archtree algorithm to predict the most promising pruned sub-model, by looking at the delta of importance w.r.t. the parent (non-pruned) model. We will now introduce our Archtree algorithm to explore the space of possible pruned sub-models.

%This importance criterion is used in two ways. First, it is the metric which determines channels to remove \textit{inside} a channel group, with the convention that we always remove the least important channel first. Second, by looking at the delta of importance w.r.t. the parent, non-pruned model, we can predict which pruned sub-model is more likely to be as performant as its parent model. However, importance alone cannot dictate how we choose $\Cc'$, \textit{i.e.} the number of channels in each channel group of the pruned sub-model. For this, we use latency, which we bnchmark on-the-fly on the target hardware. To explore the space of models to bnchmark, we propose the Archtree algorithm. \kevin{je simplifierais le paragraphe en disant juste que nous allons voir dans la section suivante, comment ce critère d'importance sera utilisé au sein de l'algo archtree pour à la fois selectionner les channels à dégager au sein d'un groupe et pour ranker les sous modèles générer à partir}

\subsection{Archtree}\label{ssec:archtree}

\begin{figure}[!t]
    \centering
    \includegraphics[width=1\linewidth]{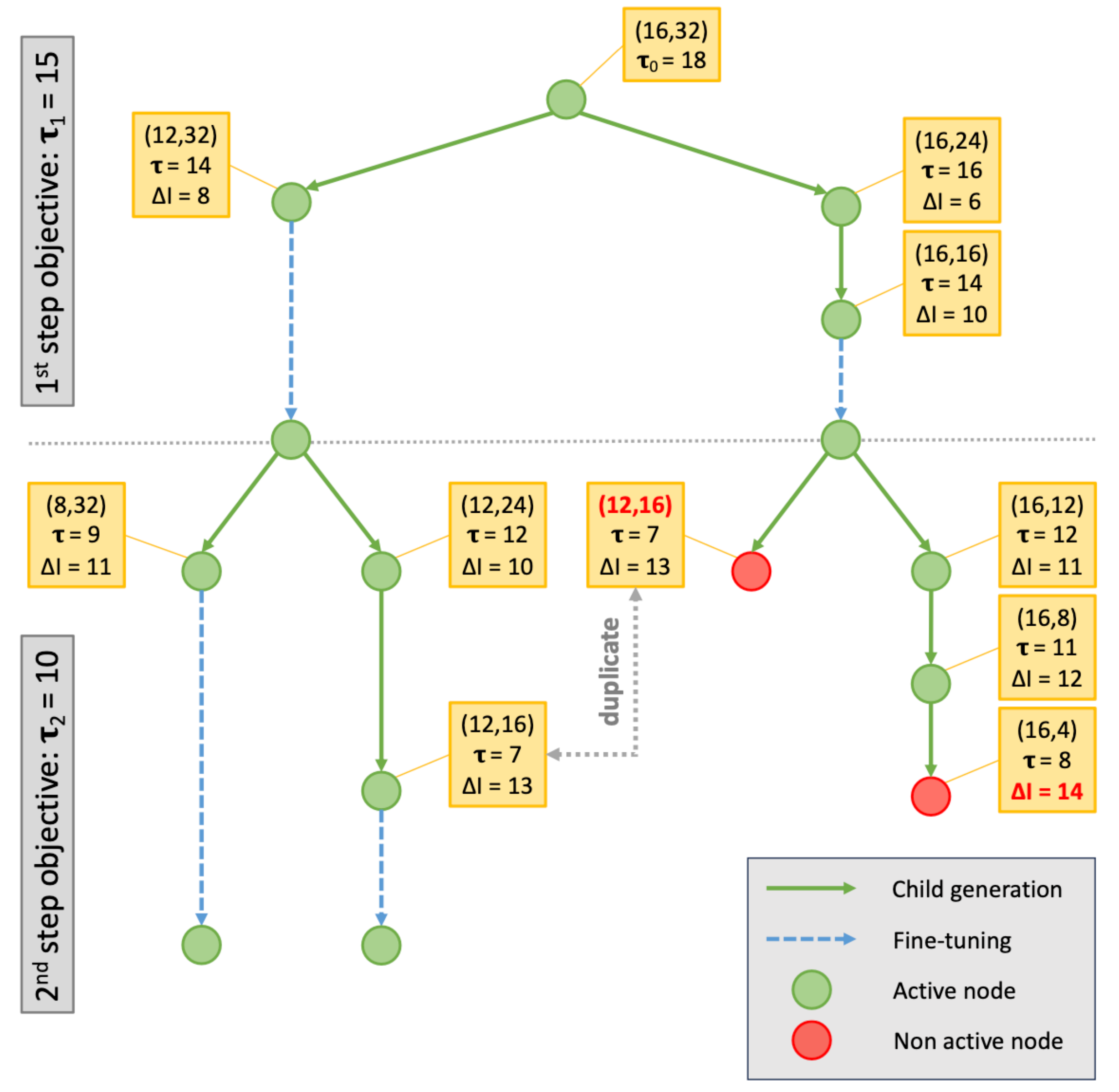}
    \caption{Archtree for a model with 2 channel groups, with 2 pruning steps shown and $A = 2$. Plain branches denote children generation, dotted branches represent fine-tuning.}
    \label{fig:archtree}
\end{figure}

In this section, we explain the proposed Archtree method, through a detailed description of its steps. These steps are also expressed in Algorithm \ref{ArchtreeAlgorithm} and illustrated in an example in Figure \ref{fig:archtree}.

Consider a model $M$  with latency $\tau_0$, called the root model, and associated to a node with signature $\Cc = \left( \Cc_1, \dots, \Cc_N \right)$. We want to prune it into a new model with latency below $\tau_s < \tau_0$. This pruning can be broken down into $s$ steps (a hyperparameter of the method), each of these involving a step-wise pruning rate objective $\tau_i$, for which we use uniform latency scheduling by setting: 
\begin{equation}
    \tau_i = \frac{ \left( s - i \right) \tau_0 + i \tau_s }{s}
\end{equation}
with $i \in [\![ 1, s ]\!]$ (gray boxes in Figure \ref{fig:archtree}).

\paragraph{Archtree initialization:} the set of alive nodes $\Aa$ is initialized with the root node only. Throughout the $s$ steps of the pruning phase, we keep a maximum of $A >= 1$ alive nodes in parallel. This is equivalent to applying beam search to the possible candidate pruned networks. This setup allows for more exploration of the search space, as compared to the greedy search in \cite{halp}, where each pruning step only keeps one pruned model. $A$ is a crucial hyperparameter of our method: the larger $A$ is, the more exhaustive the exploration of the search space becomes, at the expense of more computational load. The setting of this hyperparameter will be discussed in the experiments (\ref{exp:ablations}), but the main takeaway is that setting $A$ to a large enough size (e.g. 3 for one of our experimental setup) allows for sufficient exploration of the search space. Beyond that value, increasing $A$ offers diminishing returns. 

%Therefore, Archtree explores a larger part of the search space than HALP and offers more candidate models to finetune in one run, as long as $A \geq 1$. The impact of this effect grows with $A$. Setting $A$ to its limit of $N^s$ (no node is deactivated) we get a full exploration of the search space, up to the choice of the exploration step. \remi{TODO: check that with the table} However, we can see in table \ref{tab:active} that the performance of Archtree stops growing after a certain value of $A$, negligible compared to $N^s$. This means that setting $A$ to a high enough value is sufficient to explore the relevant part of the search space, where latency is low and accuracy high. \edouard{alors ca peut le faire mais en general on evite d'avoir des exps dans la methodo :)}

\paragraph{Step-wise fine-tuning and importance calculation phase:} for each active node $a \in \Aa$, the associated model $M_a$ is then fine-tuned upon a few batches from $\mathcal{D}$, typically 320 for a ResNet18 trained on ImageNet. Fine-tuning (illustrated by blue dashed arrows in Figure \ref{fig:archtree}) serves two main purposes. First, this allows us to mitigate the accuracy loss caused by the pruning. In fact, as pointed out in \citep{yvinec2022singe}, closely entwining smaller pruning and fine-tuning steps better preserves the original model accuracy, as compared to e.g. performing a whole fine-tuning after having pruned a larger number of channels. Second, as we perform small updates to the network weights to preserve the accuracy, we can measure the importance of each weight by applying equation (\ref{eq:wimp}), and of each channel group by applying all the reductions discussed in Section \ref{ssec:importance}\footnote{For the root model, we only backpropagate the loss for importance estimation, without updating the weights since this model does not require accuracy recovery}. Admittedly, doing so means that the importance computed at the beginning of a fine-tuning step concerns weights that will change during this fine-tuning. However, since this step is done on only a handful of batches at every step, we simply disregard the small weight variation. Although there is $A$ fine-tuning per step (one for each alive node) these can be parallelized to speed up the algorithm.

\paragraph{Step-wise pruning phase:} for each model $M_a ,~ a\in \Aa$, child nodes are generated by pruning one of the $N$ channel groups of the model. Specifically, for each channel group $n \in [\![ 1, N ]\!]$, we prune one channel in the group $n$ and check the latency of the new model, directly on the target hardware. This ensures that there is no latency approximation error, unlike in \cite{halp}. If the latency is above the step's latency goal $\tau_i$, we try pruning more channels in $n$. If it is below $\tau_i$, we have found a suitable child for the root and stop pruning channel group $n$. At the end of the process, we have a new child node with signature $\Cc'$ which is component-wise equal to $\Cc$, except for $\Cc'_n$ which is strictly lower than $\Cc_n$. If no such child can be found (for instance, if pruning channel group $n$ down to only 1 channel still does not match the latency goal $\tau_i$), we move on to another channel group without creating a child. We repeat this process with each channel group, yielding a maximum of $N$ children for the root, all with latency below $\tau_i$. Plus, we avoid duplicate nodes: once a node has been created on the tree, no new child can be created with the same signature $\Cc$.

For each of these children, using the channel-wise importance values computed during the fine-tuning phase, we compute the sum of the importance (as defined in Section \ref{ssec:importance}) among the pruned channels: this defines the loss $\Delta \mathcal{I}$ of the child model. The lower this loss, the more likely the child is to perform well. We then remove node $a$ from the active node list, and only keep the $A$ nodes with minimal loss among the child nodes.

Last but not least, after the $s$ pruning steps (with entwined fine-tuning steps) are completed, a global fine-tuning phase is performed on each of the $A$ pruned sub-models. As for the step-wise fine-tuning phase and pruning phase, this global fine-tuning phase can also be parallelized over $A$ machines, making its time cost independent of $A$.

\subsection{Efficient Archtree Exploration} \label{ssec:efficient}

As we go down the Archtree, the pruned models have progressively lower latency. Besides, since latency is measured on the fly, we know for sure that at pruning step $s$ all models have a latency below $\tau_s$. This is guaranteed because no proxy is used to compute the latency, contrary to prior work \cite{halp}. However, latency benchmarking, i.e. on-the-fly latency measurement, takes time. At every level on the Archtree, the number of such benchmarks is $\mathcal{O}\left( A \sum_{n=1}^N \Cc _n \right)$. This proves more challenging when pruning models that are deeper (large $N$) or wider (large $\Cc_n$). In order to reduce this time cost, we propose several mechanisms.

\begin{figure}[!t]
    \centering
    \includegraphics[width=\linewidth]{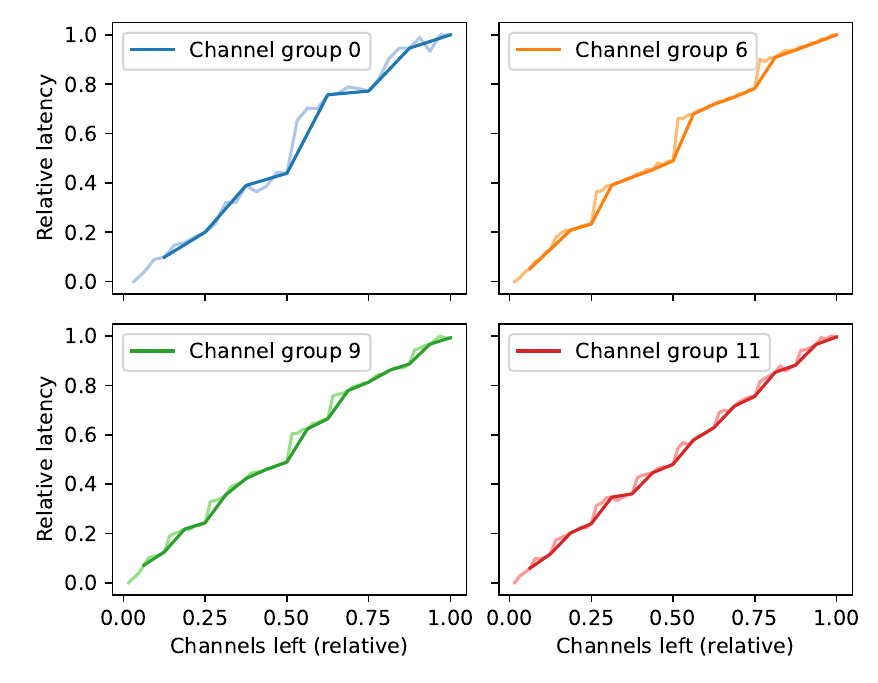}
    \caption{Relative latency of four channel groups depending on the fraction of channels left. The bold curve is the latency explored by the Archtree (due to the adaptive exploration step), while the light curve gives a more general impression of the latency curve. Latency curves are drawn by pruning a channel group inside a ResNet18.}
    \label{fig:latency_curve}
\end{figure}

\paragraph{Exploration step size} Consider we are generating a child by pruning channel group $n$. If we remove one channel at a time, there are at most $\Cc_n$ latency measures to do. However, if we instead consider removing $\delta > 1$ channels at a time, the maximum number of latency benchmarks falls to $\lceil \frac{\Cc_k}{\delta} \rceil$. This gain comes at the price of exploring the space of possible sub-models less finely. In order to find a reasonable trade-off between speed and granularity, we make $\delta$ a function of $\Cc_k$: 
\begin{equation}
     \delta \left( \Cc_k \right) = 2^{ \lceil log_2(\sqrt{\Cc_k}) \rceil }
\end{equation}
This formula ensures that $\delta \left( \Cc_k \right)$ is always a power of $2$, which can prove useful for memory-alignment reasons. Indeed, on both hardware we tested on, (e.g. a STM32 board and a GTX2070 GPU), latency often decreases when channel groups have a size that is a power of $2$. Additionally, it means the maximum number of latency benchmarks is roughly $\sqrt{\Cc_k}$, which makes the exploration manageable in our use cases. If we want to guarantee a finer exploration, we can trade the square root for a logarithm, making the step smaller: $\delta' \left( \Cc_k \right) = \mathcal{O} \left( \log \left( \Cc_k \right) \right)$. As shown in figure \ref{fig:latency_curve}, no significant pattern in the channels-to-latency curve is missed when removing more than 1 channel at a time. Furthermore, we can see that latency curves have a staircase-like aspect, albeit with slanted steps. \citet{halp} prunes channels according to this step-size, always landing at the beginning of a new step (\textit{i.e.} points after a sudden latency dip). Hence, HALP never considers the points in the middle of a step, but Archtree, thanks to a lower step-size, can. Thus, Archtree explores a more granular search space than HALP does, which can lead to better results (Tables \ref{tab:comp8} and \ref{tab:comp18}).

\begin{figure}[!t]
    \centering
    \includegraphics[width=\linewidth]{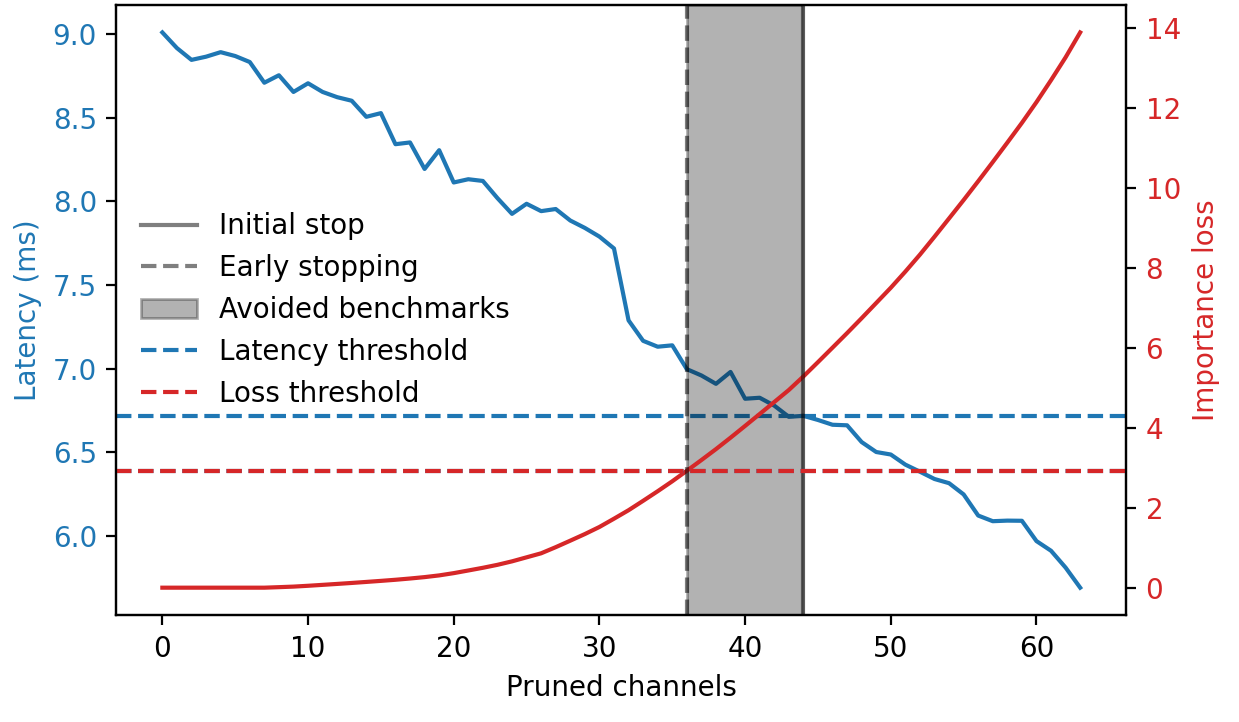}
    \caption{Loss and latency curves when pruning a channel group. The importance-based early stopping mechanism is illustrated in gray: the greyed area corresponds to latency benchmarks avoided because the maximal loss threshold was crossed before the minimal latency threshold.}
    \label{fig:early_stopping}
\end{figure}

\paragraph{Importance-based early stopping during latency benchmarking} Suppose we are in the process of generating children, and that $A$ children models $\left( M_1, \dots, M_A \right)$ have already been generated, and their importance loss $\Delta \mathcal{I}_1, \dots, \Delta \mathcal{I}_A$ have been estimated. When generating a new child, we progressively prune channels in order to reduce the child latency, hence increasing the importance loss $\Delta \mathcal{I}$ of the child. If at some point, $\Delta \mathcal{I} \geq \max_{1 \leq a \leq A} (\Delta \mathcal{I}_a)$ then we know for sure that the child will not be among the $A$ nodes with the minimal loss. In such a case, we can stop pruning channels and discard the child. This importance-based early stopping mechanism (figure \ref{fig:early_stopping}) does not affect the Archtree performance and ensures that no useless latency benchmarking is done when generating new children.

\paragraph{Caching mechanism} When we benchmark the latency of a model with channel group vector $\Cc$ and latency $\tau$ on a given target hardware, we save the mapping $\Cc \rightarrow \tau$ for later. If we need to benchmark a model with a similar $\Cc$, instead of doing the actual latency benchmarking, we return $\tau$. The time-cost of benchmarking is replaced by the low memory-cost of the memorized mapping. This is especially useful with repeated runs, for instance when tuning hyperparameters like $A$ or $s$, or when changing the latency goal $\tau_s$. 

\section{Experiments}

\subsection{Implementation Details}
All experiments were conducted using Python 3.11, PyTorch 2.0 and CUDA 11.7. Mainly, we showcase the performance of Archtree and compare it to HALP on two different testbeds to show that our method establishes a new state of the art on both edge devices and consumer GPUs. First, with a STM32 board \cite{stm32} as the target hardware, using the ResNet8 architecture from the tinyML challenge \cite{tinyml} (Since ResNet18 could not fit on this device) trained on CIFAR10 \cite{CIFAR10}. Latency measurement was done using ST's inference engine (the software which runs inference in the fastest way possible) and profiler (which measures the latency of inference). Second, using a ResNet18 trained on ImageNet \cite{imagenet} with inference on a GTX2070 Nvidia GPU. Here, latency measurements were made using PyTorch's bindings of CUDA events. To reduce the impact of noise, the reported latency was averaged over $10^4$ iterations after $10^3$ iterations of warm up. This is a lengthy benchmark, so during exploration, we measure latency by taking the median over 300 iterations, after 100 warm up iterations.
 %Since ResNet18 could not fit on to the STM32 board, we were obligated to use a ResNet8 for this device. However, there was no point to running the same model on the GTX2070, as the main latency contribution came from the sequentiality of layers rather than the size of their weights. Hence, we opted for a larger model on the GPU, ResNet18, which opened up the door for a more challenging dataset, ImageNet.

When pruning ResNet18, we use $s=30$ pruning steps, each with 320 batches of fine-tuning with batch-size 32 and SGD optimizer at $0.1$ learning rate. After all pruning steps are done, we perform a final fine-tuning phase where we use the SGD optimizer for 30 epochs with 0.01 learning rate. The learning rate schedule is 1 epoch of warm-up (or 2, for pruning rates 80 and 90\% of Archtree) followed by cosine decay. We keep $A = 3$ nodes alive during Archtree exploration.
For ResNet8, since the model is smaller, we allow for $A=6$ nodes during exploration. We prune over $s=7$ steps, each with 500 batches of fine-tuning with batch-size 32 and SGD optimizer at $0.1$ learning rate. After the pruning phase, fine-tuning is done over $23$ epochs with the same setup as for the ResNet18.

To ensure a fair test between Archtree and HALP, we set their common hyperparameters to the same value, including pruning steps and training hyperparameters.

\subsection{Ablation studies}\label{exp:ablations}

\paragraph{Importance reductions:} in subsection \ref{ssec:importance} we proposed a three steps reduction to go from per-weight importance tensors to per channel group importance vectors. We defined three operators, $\Rr_s ,~ \Rr_n ,~ \Rr_c$ which can either be a sum ($\Sigma$), a mean (Avg.) or an infinite norm ($L_\infty$). To find which combination of operators led to the best Archtree performance, we pruned a ResNet18 for on a GTX2070. The latency pruning goal was $0.7$ with $A=3$.

\begin{table}[!t]
\vskip 0.15in
\begin{center}
\caption{Impact of reduction on Archtree's final performance.} \label{tab:importance}

\begin{tabular}{|c|c|c|c|c|c|}
\hline
$\Rr_s$ & $\Rr_n$ & $\Rr_c$ & Validation accuracy  \\
\hline
\multicolumn{3}{|c|}{Original model} & 69.758 \\
\hline
$\Sigma$   & $L_\infty$ & $\Sigma$   & \textbf{69.174}  \\
$\Sigma$   & $\Sigma$   & $\Sigma$   & 68.790 \\
Avg.       & $L_\infty$ & $\Sigma$   & 68.074 \\
Avg.       & $\Sigma$   & $\Sigma$   & 67.578 \\
$L_\infty$ & $\Sigma$   & Avg.       & 67.534 \\
$\Sigma$   & Avg.       & Avg.       & 66.928 \\
Avg.       & $L_\infty$ & $L_\infty$ & 62.774 \\
\hline

\end{tabular}
\end{center}
\vskip -0.1in
\end{table}

The results are reported in table \ref{tab:importance}: the best results are obtained when $\Rr_c = \Sigma$: indeed, $\Sigma$ is the only reduction that takes into account the number of layers impacted by the pruning of a channel group (The more layers are pruned, the larger the importance loss). For neural reduction $\Rr_n$, using $L_\infty$ over $\Sigma$ yields better results, as comparing lines 1 and 2 or 3 and 4 reveals. Regarding the spatial reduction $\Rr_s$, $\Sigma$ performs best, as the first four lines show. The last 3 lines illustrate why all 27 combinations were not tested: some reduction choice leads to poor final results, as can be expected, like $\Rr_c \neq \Sigma$. Another observation is that the naive approach $\left( \Sigma, \Sigma, \Sigma \right)$ (similar to that of previous work \cite{halp} is outperformed by the proposed $\left( \Sigma, L_\infty, \Sigma \right)$.
Consequently, for the remainder of this article, we will assume the best combination overall is $\left( \Rr_s, \Rr_n, \Rr_c \right) = \left( \Sigma, L_\infty, \Sigma \right)$.

\begin{figure}[!t]
    \centering
    \includegraphics[width=\linewidth]{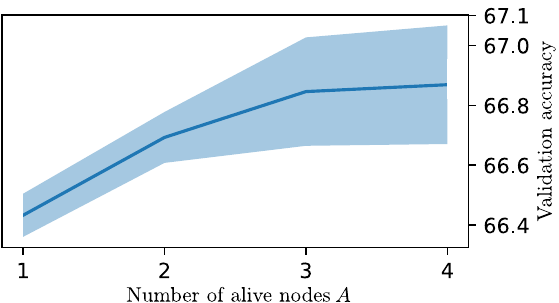}
    \caption{Impact of the number of alive nodes $A$. The plain line indicates the average accuracy and the filled area the standard deviations over 5 runs.}
    \label{fig:ablationA}
\end{figure}

\paragraph{Alive nodes:} Figure \ref{fig:ablationA} shows the variation of accuracy on a ResNet-18 pruned with a latency goal of $0.5$ for a GTX2070 GPU, when varying the number of active nodes $A$. The average accuracy steadily increases when $A$ goes from 1 to 3 active nodes at a time, and begins to reach diminishing returns for $A \geq 3$. Hence, in what follows, we will use $A=3$ for all the experiments on the GTX 2070. When changing hardware, we can start with $A=3$ and gradually increase that value until a plateau is reached, as was done on STM32. Thanks to the caching mechanism, those repeated runs are quite inexpensive compared to the first run.

%\begin{table}[!t]
%\vskip 0.15in
%\begin{center}
%\caption{Impact of the number of alive nodes $A$. \arnaud{virer la 3e colonne?}}
%\label{tab:active}
%\begin{tabular}{|c|c|c|}

%\hline
%Alive nodes & Validation accuracy & Relative latency \\
%\hline
%1 & 66.433 $\pm$ 0.072 & \\ % Runs: 0/3: 77 OK - 78 OK - 79 OK
%2 & 66.693 $\pm$ 0.085 & \\ % Runs: 1/3: 80 OK - 82 OK - 83 OK
%3 & 66.846 ± 0.181 & 0.5 ± 0.0 \\ % Runs: 5/5 - All ok
%4 & 66.869 $\pm$ 0.198 & \\ % Runs: 5/5 - 72 OK - 71 OK - 70 OK - 69 OK - 56 OK 
%5 & 66.785 $\pm$ 0.168 & \\ % Runs: 5/5 - 64 OK - 66 OK - 67 OK - 68 OK - 73 OK
%6 & & \\ % Runs: 0/3 
%7 & 66.162 $\pm$ 1.292 & \\ % Runs: 3/3 - 74 OK - 75 OK - 76 OK
%\hline
%Original model & 69.758 & 1.0 \\
%\hline

%\end{tabular}
%\end{center}
%\vskip -0.1in
%\end{table}

\subsection{Efficient exploration validation}

\begin{figure*}[t!]
    \centering
    \includegraphics[width=\linewidth]{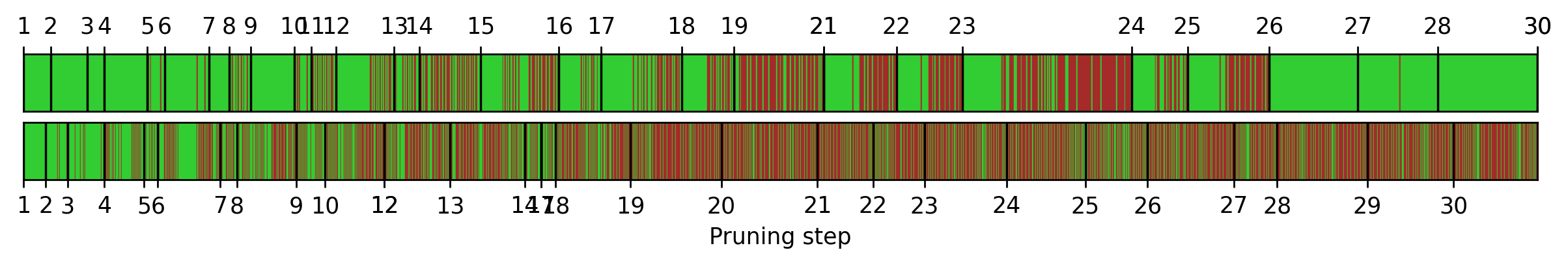}
    \caption{Illustration of latency benchmarks done throughout the pruning steps (x-axis) during the second run of a repeated Archtree run (setup \textbf{(A)}, top bar, 71.7\% hit rate) and a second run where the pruning goal changed (setup \textbf{(B)}, bottom bar, 43.96\% hit rate). These benchmarks are cache hits (green - latency measure was already in memory) or misses (red - latency was actually measured).}
    \label{fig:caching}
\end{figure*}

To reduce the time cost of descending the Archtree, we have proposed three mechanisms: adaptive exploration step size, importance-based early stopping and caching. They all reduce the number of latency benchmarks, as it is the main cost of the exploration. To validate those choices, we look at the number of benchmarks of the search with each mechanism on or off. For each mechanism, the results are gathered when pruning a ResNet18 \cite{resnet_arxiv} on GTX2070, with $A=3$ and pruning rate $0.5$ (remove 50\% of the latency).

\paragraph{Exploration step size} To measure the effect of having an adaptive exploration step size, we note how many channels there are in each group before and after each child is generated. We then compute the number of latency benchmarks done and the potential number of such benchmarks if the step was set to 1. Here, we assume that there are no large irregularities in the channel-to-latency curve, which is what we observed e.g. on GPU, as shown in figure \ref{fig:latency_curve}. This can be false in edge cases or because of very noisy measures. To compensate for this, we will assume the worst case for the exploration step, which is that the penultimate benchmark is always 1 channel short of the latency threshold. As an example, if the exploration step is $8$ and exploration goes: $32, 24, 16$ channels, the worst case is that the actual latency threshold was at $23$ channels.\\
Using this approximation, we infer the gain from exploration step size over 5 runs. We find that without an adaptive exploration step size ($\delta = 1$) the number of latency benchmarks is $10.97 \pm 0.18$ times larger. Hence, although the exploration step is the only mechanism in this section to reduce the size of the search space, it almost multiplies the exploration speed of the Archtree by 11.

\paragraph{Importance-based early stopping} After deactivating the importance-based early stopping mechanism, the number of benchmarks in the Archtree descent goes from $1928$ to $3763$. The proposed importance-based early stopping never changes the result of the Archtree, so in this run it prevented $51.2\%$ of latency benchmarks at no cost. This test was conducted with adaptive exploration step size on. 

\paragraph{Caching} We estimate the impact of caching with two experimental setups. Setup \textbf{(A)} comprises two repeated runs, a situation that can arise because we want to generate more candidate pruned models or when tuning hyperparameters. Setup \textbf{(B)} is a run with pruning rate 0.7 followed by a run with pruning rate 0.5, which corresponds to the situation where we progressively increase the pruning rate. In both setups, we look at the number of cache hits and misses. A cache hit occurs when the model benchmarked for latency has a channel group vector $\Cc$ that has already been benchmarked. A cache miss is the opposite.\\
For setup \textbf{(A)}, the repeated run has $1332$ benchmarks, $71.7\%$ of them being cache hits. Thus, the time spent actually benchmarking in the repeated run is less than $30\%$ of what it was in the original run. Furthermore, we can look at the pattern (Figure \ref{fig:caching}) of cache hits (green) and misses (red) to get insight into the repeated Archtree descent. We observe that at first (left portion of the plot), there are almost only hits, meaning that the descent through the tree is almost fully consistent across both runs. Then, there are more misses, because the two descents diverge (due to randomization of the fine-tuning process). The diverging branch starts small and grows larger and larger, leading to progressively more misses. Nevertheless, in the end (right side of the plot), there are only hits: the two runs only diverge in the middle of the descent, but converge to the same leaves. This can be interpreted as the Archtree finding the same low-latency, high-accuracy models twice, hence assessing the stability of the proposed method.\\
For setup \textbf{(B)}, the second run has a hit rate of $43.96\%$ over $1986$ latency benchmarks. While this is expectedly lower than for the repeated runs (because of the change in latency goal) we still cut almost half of the latency-measurement time without impacting the Archtree's result. In this run, hits are more common in the beginning, meaning the Archtree starts roughly the same with both pruning goals, and then hits and misses are spread uniformly throughout the rest of the run.  

\begin{table*}[!t]
\begin{center}
\caption{Archtree and HALP on ResNet8 - STM32. Original model has 87.190 accuracy and latency of 33.383ms.}
\label{tab:comp8}
\begin{tabular}{|c||c|c|c||c|c|c|}
\hline
& \multicolumn{3}{c||}{Archtree} & \multicolumn{3}{c|}{HALP} \\
\hline
Prune rate & Val. acc & Relative lat. & Params & Val. acc & Relative lat. & Params \\ 
\hline
0.9 & 87.51 $\pm$ 0.067  & \textbf{0.89} $\pm$ 0.0  & 70356 & \textbf{88.18}  $\pm$ 0.02  & 0.91 $\pm$ 0.0  & 74336 \\ 
0.8 & \textbf{85.968} $\pm$ 0.259 & \textbf{0.8 } $\pm$ 0.0  & 63106 & 85.41  $\pm$ 0.04  & 0.79 $\pm$ 0.01 & 65266 \\
0.7 & \textbf{84.912} $\pm$ 0.283 & \textbf{0.69} $\pm$ 0.01 & 59743 & 84.505 $\pm$ 0.075 & 0.72 $\pm$ 0.0  & 64310 \\ 
0.5 & \textbf{82.008} $\pm$ 0.279 & \textbf{0.49} $\pm$ 0.01 & 42683 & 76.435 $\pm$ 0.085 & 0.51 $\pm$ 0.0  & 43052 \\ 
0.3 & \textbf{76.410} $\pm$ 0.209 & \textbf{0.3 } $\pm$ 0.0  & 14842 & 50.9   $\pm$ 4.89  & 0.32 $\pm$ 0.0  & 4106 \\ 
\hline
\end{tabular}
\end{center}
\vskip -0.1in
\end{table*}

\begin{table*}[!t]
\vskip 0.1in
\begin{center}
\caption{Archtree and HALP on ResNet18 - GTX2070. Original model has 69.758 accuracy and latency of 15.139ms.}
\label{tab:comp18}
\begin{tabular}{|c||c|c|c||c|c|c|}
\hline
& \multicolumn{3}{c||}{Archtree} & \multicolumn{3}{c|}{HALP} \\
\hline
Prune rate & Val. acc & Relative lat. & Params ($\times 10^6$) & Val. acc & Relative lat. & Params ($\times 10^6$) \\ 
\hline
0.9 & \textbf{69.552} $\pm$ 0.045 & \textbf{0.88} $\pm$ 0.02 & 11.597 & 69.327 $\pm$ 0.136 & 0.89 $\pm$ 0.01 & 11.196 \\ 
0.8 & \textbf{69.308} $\pm$ 0.041 & \textbf{0.79} $\pm$ 0.01 & 11.537 & 68.939 $\pm$ 0.08  & 0.85 $\pm$ 0.0  & 9.405  \\
0.7 & \textbf{68.970} $\pm$ 0.133 & \textbf{0.7 } $\pm$ 0.01  & 11.404 & 67.807 $\pm$ 0.032 & 0.78 $\pm$ 0.01 & 6.381  \\ 
0.5 & \textbf{66.846} $\pm$ 0.181 & \textbf{0.5 } $\pm$ 0.0   & 10.859 & 66.411 $\pm$ 0.31  & 0.53 $\pm$ 0.01 & 6.499  \\ % 15-19
0.3 & \textbf{61.982} $\pm$ 0.194 & \textbf{0.3 } $\pm$ 0.0   & 6.461  & 55.288 $\pm$ 0.448 & 0.36 $\pm$ 0.0  & 1.334  \\ % 20-24
\hline
\end{tabular}
\end{center}
\vskip -0.1in
\end{table*}

\subsection{Comparison to State-of-the-Art Pruning methods}
The current state-of-the-art in structured pruning, such as SInGE \cite{yvinec2022singe}, focuses on removing as many parameters as possible with no explicit consideration for the final latency. As explained in Section \ref{sec:intro}, this may however lead to suboptimal solutions in terms of latency: for instance, on ResNet 8, SInGE reduces the number of parameters by 66.6\% while reaching an accuracy of 88\%. However, in practice, this only results in a 6.54\% latency speed-up. On the flip side, the proposed Archtree method can achieve up to 3 times higher latency improvements over SInGE while removing significantly fewer parameters (23.11\%). This highlights the importance of latency-driven pruning for efficient inference. \\
Consequently, in the remainder of this section, we will focus on the comparison between Archtree and the state-of-the-art in latency-driven structured pruning \cite{halp}.

\paragraph{ResNet8} In Table \ref{tab:comp8}, we highlight the performance of Archtree as compared to the state-of-the-art latency-driven pruning technique HALP \cite{halp} on a low footprint model (ResNet8) on a low power chip (STM32). Our observations are three-fold. First, For any target latency (pruning rate), Archtree systematically offers the desired latency or a slightly faster result, contrary to HALP which does not always provide results very close to the target latency. Second, the pace at which parameters are removed as the latency target decreases is very unstable for HALP. On the flip side, Archtree allows a much more steady, and stable profile for parameter removal as the target latency decreases. This leads to our third observation, where Archtree keeps more parameters (every rate except 0.9) it offers a higher accuracy at a lower latency. This is blatant in the high pruning regimen, e.g. for a 0.7 latency pruning rate, where Archtree outperforms HALP by 25.51 points.

\paragraph{ResNet18} As shown in Table \ref{tab:comp18}, the proposed Archtree outperforms HALP for each of the 5 target latencies tested, both in terms of effective latency and accuracy. When it comes to latency, Archtree always satisfies the pruning rate we set, sometimes going a little over the limit, especially for lower pruning rate regimen (0.9 and 0.8). As for accuracy, Archtree outperforms HALP, cutting the drop in accuracy by 2 in some cases where it is already low (pruning rate 0.8 or 0.7). Furthermore, here again, Archtree shines for larger pruning rates: with a pruning rate of 0.3, Archtree achieves a drop in accuracy of only 7.78 points as compared to the 14.47 points drop of HALP. In this context, HALP is also far from reaching the required 0.3 latency goal. This can be explained by looking at the number of parameters of the models: Archtree finds a better parameters-to-latency trade-off and keeps almost five times more parameters. 

%A global comparison of the two techniques is offered by figure \ref{fig:pruningComp} (left side), showcasing that Archtree is more accurate and faster than HALP for all pruning rates. The only one for which it is not clear is the lowest pruning rate (90\%) where Archtree performs only a tiny bit better. Also, for the half pruning rate (50\%), the space between curves is lower than what we might expect. This might be because HALP's latency-to-accuracy curve is decreases slowly for low pruning rates (90-50\%) and then suddenly dips (30\%) whereas Archtree's curve decreases in a more regular fashion. \arnaud{Par ailleurs les conclusions sur les setup experimentaux sont je trouve assez petit bras!}

Generally speaking, Archtree outperforms HALP in terms of accuracy in nearly every scenario, and particularly for large pruning rates, which we attribute to a better exploration of the search space of the pruned sub-models. Furthermore, on-the-fly \textit{in situ} latency estimation allows to better estimate the impact of removing specific channels on the latency of the final model: this, in turn, allows to more closely fit the latency budget while, coincidentally, enabling a much steadier profile for the parameter number \textit{vs.} pruning rate curve and, as such, an overall more stable algorithm behavior. 

This shows the interest of our method for latency-aware structured pruning of DNNs. Below, we provide concluding remarks and pinpoint some limitations of the proposed approach, that shall guide future research.

%the only one where it does not is the lowest (90\%), where Archtree falls short of HALP by 0.7 points of accuracy. However, HALP models have benefited from a grid-search on its hyperparameters for the fine-tuning phase, whereas Archtree did not. Additionally, Archtree has a speedup of 89\% (below the goal) where HALP shows a speedup of 91\% (above the goal). 

%However, as for the rest of the pruning rates, Archtree shows better results than HALP, again especially for large pruning rates, where HALP's performance severely dips. This is apparent in figure \ref{fig:pruningComp} (right side) and the number of parameters in each model: once again, Archtree finds a better parameters-to-latency trade-off.

\section{Conclusion}
In this paper, we showed the limitations of existing structured pruning methods when it comes to direct translation into inference latency gains. Furthermore, we pinpointed the limitations of the most successful latency-aware approach \cite{halp}, namely the fact that only a single candidate pruned model is considered at a time, and the fact that latency is only estimated off-line using a lookup table and in a layer-wise manner, not accounting for problems linked with serialization, parallelization, or hardware-rooted side effects such as memory transfers or caching. The former shortcoming limits its accuracy, while the latter leads to poor estimation of the actual latency of the resulting pruned model w.r.t. the latency goal.

Consequently, in this paper, we introduced a novel approach to latency-driven structured pruning, dubbed Archtree. Archtree involves tree-structured search within the pruned sub-model space, effectively maintaining multiple candidates at a time. Furthermore, Archtree uses on-the-fly latency measurement in order to systematically achieve a lower or equal latency as targeted. Experimentally, Archtree outperforms previous structured pruning techniques in terms of accuracy \textit{v.s.} speed trade-offs on multiple convolutional neural networks on both low-power edge devices and consumer GPUs.

\paragraph{Limitations and future works:} nevertheless, the proposed method has room for improvement. As such, Archtree would benefit from using better importance criteria, such as the one proposed in \cite{yvinec2022singe}, to better identify promising sub-models. Additionally, it would be useful to reduce either the number or the length of pruning steps, to make for a faster algorithm. Finally, the proposed Archtree could be used to conduct pruning on more recent DNN architectures, such as transformers \cite{dosovitskiy2021vit}. Transformers prove more accurate than convolutional-based models and are well-suited for latency profiling, as they are composed of repeating and sequential blocks. By profiling the latency of one block and generalizing the measure onto others, we could in theory replace some of the costly latency benchmarks with approximation, while still referring to on-the-fly benchmarking for validation of these estimations.

\bibliography{paper}
\bibliographystyle{style}

\end{document}